%% file: main.tex
\documentclass[conference]{IEEEtran}
\IEEEoverridecommandlockouts
\usepackage{cite}
\usepackage{amsmath,amssymb,amsfonts}
\usepackage{algorithmic}
\usepackage{mathtools}
\usepackage{graphicx}
\usepackage{textcomp}
\usepackage[pdftex,dvipsnames]{xcolor} 
\def\BibTeX{{\rm B\kern-.05em{\sc i\kern-.025em b}\kern-.08em
    T\kern-.1667em\lower.7ex\hbox{E}\kern-.125emX}}

\newcommand{\insertref}[1]{}
\newcommand{\explainindetail}[1]{}
\newcommand{\improvement}[1]{}

\newcommand{\permission}{%
\begin{figure}[b]{%
\footnotesize
{\includegraphics[scale=0.7]{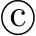}}
2019 IEEE. Personal use of this material is permitted. Permission from IEEE must be obtained for all other uses, in any current or future media, including reprinting/republishing this material for advertising or promotional purposes, creating new collective works, for resale or redistribution to servers or lists, or reuse of any copyrighted component of this work in other works.
}
\end{figure}
}

\begin{document}

\title{The Receptive Field as a Regularizer \\ in Deep Convolutional Neural Networks \\ for Acoustic Scene Classification
}

\author{\IEEEauthorblockN{Khaled Koutini$^1$, Hamid Eghbal-zadeh$^{1,2}$, Matthias Dorfer$^1$, and Gerhard Widmer$^{1,2}$}
% \IEEEauthorblockA{$^1$\textit{Institute of Computational Perception}, \\
$^1$\textit{Institute of Computational Perception\&} % \\
$^2$\textit{LIT Artificial Intelligence Lab,} \\
\textit{Johannes Kepler University Linz, Austria}

}

\maketitle

\begin{abstract}
\input{abstract}
\end{abstract}


\section{Introduction and Related Work}
\label{sec_intro}
\permission
\input{sec_intro}

\section{The Effect of the Receptive Field}
\label{sec_rf}
\input{sec_rf}

\section{Experimental Setup}
\label{sec_exp}

\input{sec_exp}

\section{Experiments}
\label{sec_res}
\input{sec_res}

\section{Discussion}
\label{sec_disc}
\input{sec_disc}

\section{Conclusion}
\label{sec_con}
\input{sec_con}

\section*{Acknowledgment}
This work has been supported by the LCM – K2 Center within the framework of the Austrian
COMET-K2 program.
\bibliographystyle{IEEEtran}
%\clearpage
%\footnotesize

\bibliography{zoteroref_noedit,ref}

\end{document}

%% file: abstract.tex
Convolutional Neural Networks (CNNs) have had great success in many machine vision as well as machine audition tasks.
% However, advances in computer vision approaches was not reflected completely in the acoustic domain. 
Many image recognition network architectures have consequently been adapted for audio processing tasks. 
However, despite some successes, the performance of many of these did not translate from the image to the audio domain.
For example, very deep architectures such as ResNet~\cite{heDeepResidualLearning2016} and DenseNet~\cite{huangDenselyConnectedConvolutional2017}, which significantly outperform VGG~\cite{simonyanVeryDeepConvolutional2014} in image recognition, do not perform better in audio processing tasks such as Acoustic Scene Classification (ASC).
% failed to outperform VGG~\cite{} based architectures for Acoustic Scene Analysis (ASA) tasks. 
In this paper, we investigate the reasons why such powerful architectures perform worse in ASC compared to simpler models (e.g., VGG). 
To this end, we analyse the receptive field (RF) of these CNNs and demonstrate the importance of the RF to the generalization capability of the  models.
Using our receptive field analysis, we adapt both ResNet and DenseNet, achieving state-of-the-art performance and eventually outperforming the VGG-based models.
We introduce systematic ways of adapting the RF in CNNs, 
and present results on three data sets that show how changing the RF over the time and frequency dimensions affects a model's performance.
% We provide empirical results demonstrating that having very small or very large receptive fields, the performance since the model decreases and fails to fit the data properly and 
% fails to fit or over-fit the training data,
%  we show the model generalizes better within a specific range of receptive field.
Our experimental results show that very small or very large RFs can cause performance degradation, but deep models can be made to generalize well by carefully choosing an appropriate RF size within a certain range.

%% file: sec_intro.tex
%!TEX root = main.tex

% \subsection{Overview}

% \subsection{Related Work}
In image processing, deep Convolutional Neural Networks (CNNs) have
revolutionized the way image recognition tasks are addressed.
A central source of their power is the ability to learn multi-level internal features and representations. While lower-level
network layers learn to detect simple features such as edges, deeper layers combine these features to detect higher-level concepts such as textures, shapes, and objects. 

The current state of the art in this domain are \textit{ResNet}~\cite{heDeepResidualLearning2016} and \textit{DenseNet}~\cite{huangDenselyConnectedConvolutional2017} variants, which outperform earlier (and less deep) \textit{VGG}-based~\cite{simonyanVeryDeepConvolutional2014} architectures by a significant margin. 
This is mainly because they address shortcomings of VGG such as the vanishing gradient. 
However, in acoustic tasks (e.g., \textit{Acoustic Scene Classification; ASC}) the state of the art is still heavily dominated by VGG-like architectures~\cite{eghbal-zadehCPJKUSubmissionsDCASE20162016,lehnerClassifyingShortAcoustic2017,DorferDCASE2018task1,SakashitaDCASE2018task1}. 

For instance, Eghbal-zadeh et al.~\cite{eghbal-zadehCPJKUSubmissionsDCASE20162016} adapted the VGG architecture taken from the computer vision domain and achieved good performance in acoustic scene classification, using spectrograms as network input.
% They fed the spectrograms extracted from audios to the VGG neural network.
Hershey et al.~\cite{hersheyCNNArchitecturesLargescale2017} compared various well-known image recognition CNN architectures on a large-scale dataset of 70M audio clips from YouTube.
They showed that on such a large dataset, very deep CNNs such as ResNet-50 can perform very well. 
%Although these rules still hold for acoustic signals to some extent~\cite{dielemanMultiscaleApproachesMusic2013}, (as we empirically show in this paper) 
However, as we will show later, training such deep architectures on smaller datasets results in heavy overfitting on the training samples.
Because of this issue, many state-of-the-art ASC systems use shallower CNN architectures~\cite{eghbal-zadehCPJKUSubmissionsDCASE20162016,hersheyCNNArchitecturesLargescale2017,lehnerClassifyingShortAcoustic2017,DorferDCASE2018task1,SakashitaDCASE2018task1,DorferDCASE2018task2,IqbalDCASE2018task2,LeeDCASE2017task4,KoutiniDCASE2018task4}. % in various acoustic analysis tasks. 
 Pons et al.
\cite{ponsExperimentingMusicallyMotivated2016} investigated the effect of filter shapes in shallow CNN architectures in music classification tasks.
They proposed to change the shape of convolutional filters in order to restrict CNNs to learn either temporal or frequency dependencies in the data.

In this paper, we %propose a new technique to 
systematically investigate the effect of restricting the Receptive Field (RF) on the time or the frequency dimensions in deep and complex CNN architectures. 
We show that for relatively smaller datasets, CNNs tend to overfit if their RF covers large areas of the spectrograms. 
Moreover, limiting the RF over the frequency dimension helps output neurons generalize better to unseen samples.
We introduce an approach to adapt deep CNN architectures for ASC by restricting their RF, since the RF grows when the depth of a CNN increases.
We apply this method on a deep DenseNet~\cite{huangDenselyConnectedConvolutional2017} and ResNet~\cite{heDeepResidualLearning2016} architecture and compare classification performance before and after this adaptation.
Furthermore, we modify a number of convolutional layers in a ResNet architecture by changing their filter shapes to obtain a specific RF. %This is crucial,
The filter \emph{shape} controls the RF of its layer over the time or frequency dimensions, while the \emph{number} of modified layers
% of layers whose filters are changed,
controls the RF of the whole network. 
This technique allows us to study how gradually increasing the RF of a CNN affects its performance.

%Our work is different than~\cite{ponsExperimentingMusicallyMotivated2016} since ...

% \subsection{Motivation}

% In this paper, we investigate the performance degrading when using vanilla deeper CNNs. Furthermore we propose modified versions of ResNet and DenseNet that reaches and outperforms SOTA VGG-like CNNs in AST.

% In this paper we show that for relatively smaller datasets, such deep architectures tend to overfit on the training data. Because of this issue, many ASC systems use more shallower CNN architectures~\cite{eghbal-zadehCPJKUSubmissionsDCASE20162016,hersheyCNNArchitecturesLargescale2017,lehnerClassifyingShortAcoustic2017,DorferDCASE2018task1,SakashitaDCASE2018task1,DorferDCASE2018task2,IqbalDCASE2018task2,LeeDCASE2017task4,KoutiniDCASE2018task4} in various acoustic analysis tasks.

% 

%% file: sec_rf.tex
%!TEX root = main.tex

\subsection{The Receptive Field in CNNs}

\begin{figure*}[htbp]
\centerline{\includegraphics[]{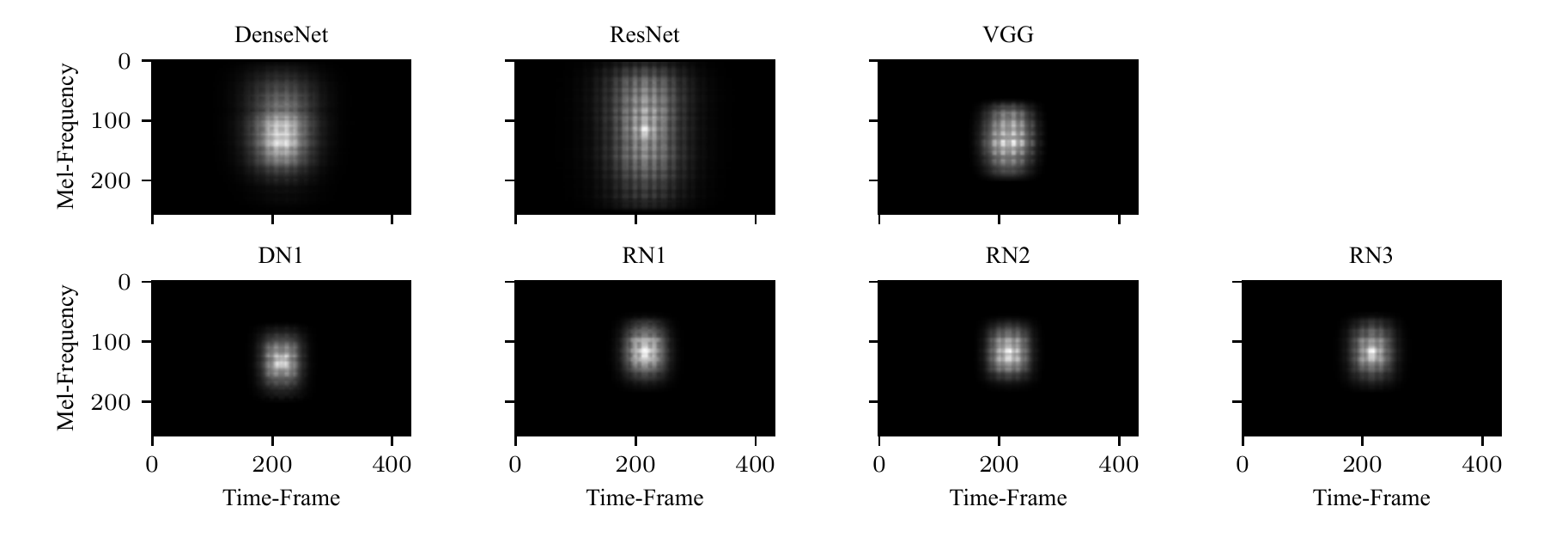}}
\caption{The Effective Receptive Field (ERF) of different CNN architectures trained on DCASE18 
(explained in Section~\ref{subsection:modifyvisionarcs}).}
\label{fig_rf}
\end{figure*}

In fully-connected layers, each neuron is affected by the whole input. In contrast, in convolutional layers each neuron has a strictly limited `field of view' (\textit{RF});
%`Receptive Field',
input values outside of this RF cannot influence the neuron's activation.
The RF in general includes input values in the spatial as well as the channels dimensions. 
However, in this paper, we will keep our focus on the spatial dimensions (the frequency and the time dimension of spectrograms, in the case of audio).

Within a convolutional layer, the spatial RF of a neuron is determined by its filter size: the bigger the filter, the more activations it can see from the previous layer. 
% However, in a CNN 
As the CNN's depth increases by stacking more convolutional layers,
% ~\cite{simonyanVeryDeepConvolutional2014,heDeepResidualLearning2016,huangDenselyConnectedConvolutional2017},
the RF of a neuron w.r.t.~the input layer -- that is, what the neuron `sees' of the input layer -- is affected by various factors such as the filter size, the stride and the dilatation of all the previous layers.
The maximum RF size can be calculated using the following equation: % \eqref{eq_max_rf}
\begin{equation}
\label{eq_max_rf}
 \begin{array}{l}
S_n=S_{n-1}*s_n \\
RF_n= RF_{n-1}+(k_n-1)*S_n
\end{array}
\end{equation}
where $s_n$, $k_n$ are stride and kernel size of layer $n$, respectively, and $S_n$, $RF_n$ are cumulative stride and RF of a unit from layer $n$ to the network input.
While the formula above gives the maximum RF each neuron has access to, a given neuron may not actually use all of it. The set of input pixels or units that effectively influence a neuron is called its \textit{Effective Receptive Field (ERF).}
Luo et al.~\cite{luoUnderstandingEffectiveReceptive2016} showed that neurons are more affected by the input pixels around the center of the RF, since these have more paths to the neuron on both the forward and the backward pass. This is important considering that in some cases the maximum RF of a network even exceeds the input dimensions.
Further, they propose a method for computing the Effective Receptive Field (ERF) of a trained model. They back-propagate a gradient signal from the output layer through the network to the input.
% \begin{figure}[htbp]
% \centerline{\includegraphics[width=0.6\columnwidth]{figs/rf0.png}}
% \caption{The effective RF of 3 different architectures explained in Section~\ref{sec_exp}.}
% \label{fig_rf}
% \end{figure}
In order to compute the ERF on models trained on audio spectrograms,
%we propose a modification to the ERF proposed by Luo et al.~\cite{luoUnderstandingEffectiveReceptive2016}. 
we follow the approach proposed by Luo et al.~\cite{luoUnderstandingEffectiveReceptive2016}.
We back-propagate a gradient signal of 1 in one output spatial pixel (i.e. before the global averaging) to the inputs.
For visualisation, we average this gradient for all the test set samples and plot the normalized average.
Figure~\ref{fig_rf} shows the ERF for different CNNs trained on the ASC task, the details of their architectures is explained in Section~\ref{subsection:modifyvisionarcs}. 
As can be seen, the ERFs can be vastly different depending on the network architecture.

\subsection{Modifying the Receptive Field of a CNN}
%As discussed in Section~\ref{sec_intro}, 
Equation~\eqref{eq_max_rf} shows that there are various ways to modify the RF of a CNN.
In this paper, we investigate how changing the RF influences the performance of a model in the task of acoustic scene classification. 
In order to achieve this goal, we follow two approaches to change the RF either by altering the filter sizes or by adding sub-sampling layers.
We detail these two approaches as follows.

% in case of lake of space this is redendant 
\subsubsection{Changing the filter sizes}
Starting from a CNN architecture %with an RF equal to the state of the art
we change the filter sizes of some layers $k_n$ from 3x3 to 1x1. One side effect of changing the filter sizes, is that it changes the number of network parameters.

\subsubsection{Changing the sub-sampling layers}
We add more max-pooling layers to increase the accumulative stride $S_n$. 
This results in changing the RF on the network, without changing the network's number of parameters.

%% file: sec_exp.tex
%!TEX root = main.tex
% what kind of experiemtn we design we mkae here to investigate.

In this section we detail the datasets, architectures and training procedure we follow in our experiments.

% (explain for what purpose we choose the dataset)

\subsection{Datasets}
We tested our hypotheses on multiple datasets with different properties, scales, and of increasing difficulty.

\subsubsection{DCASE2016}
The acoustic scenes classification task of DCASE 2016~\cite{mesarosTUTDatabaseAcoustic2016}  
is to recognise 15 possible acoustic scenes from $30$ second snippets of audio. There are $1170$ training and $390$ test examples, for a
total of 9 hours 45 minutes of training data and 3 hours 15 minutes for evaluation.
% Each audio segment can is classified to one out of 15 possible acoustic scene
% \explainindetail{maybe says why it the easiest foreshadowing the small differences in accuracy}

\subsubsection{DCASE2017}

The DCASE 2017~\cite{mesarosDCASE2017Challenge2017} training set comprises both the training and evaluation sets of DCASE 2016. The evaluation data is a set of new recordings taken at a later time. This introduces a distribution mismatch between the training and unseen test data, making for a more difficult learning task. Additionally, the audio clips are cut into 10 seconds samples totaling 4680 training samples (13 hours) and 1620 testing samples (4.5 hours). 
% Meanwhile a new evaluation set was recorded.
% This introduces a distribution mismatch between the training and unseen test data, making for a more difficult learning task.

\subsubsection{DCASE2018}

DCASE 2018 Task 1~\cite{MesarosDCASE2018T1} consists of around 17 hours of audio for training (6122 10-second clips) and 7 hours for evaluation (2518 10-second clips)\insertref{double check}. The data was recorded in more diverse locations compared to the previous challenges. % and is thus more difficult.
Each recording belongs to one out of 10 possible classes.

\subsection{Network Architectures}
\label{sec_archs}
For our experimental investigations, we used three different CNN architectures which were successful in the computer vision domain.

\subsubsection{VGG}
Architectures similar to VGG~\cite{simonyanVeryDeepConvolutional2014} are very popular in the DCASE community and the audio processing community in general. This is mainly due to their good performance in tasks such as ASC~\cite{eghbal-zadehCPJKUSubmissionsDCASE20162016,lehnerClassifyingShortAcoustic2017,DorferDCASE2018task1,SakashitaDCASE2018task1}, audio tagging~\cite{DorferDCASE2018task2,IqbalDCASE2018task2}, and event detection~\cite{LeeDCASE2017task4,KoutiniDCASE2018task4}.
%We note that these approaches generally use networks similar to VGG-13 (a VGG variant with 13 layers). While deeper variants of VGG perform worse on audio signals unlike the case of ImageNet~\cite{dengImagenetLargescaleHierarchical2009}.\gerhard{I don't understand this sentence.}
We make an effort to explain this good performance by the fact that deeper CNNs have a bigger RF as explained in Section~\ref{sec_disc}.

\subsubsection{ResNet}
\label{sec_resnet}
ResNet~\cite{heDeepResidualLearning2016} 
adds residual connections between convolutional layers, this counters the vanishing gradient problem of deep CNNs.
It outperforms VGG in computer vision tasks. In contrast, top performing systems in acoustic scene classification DCASE challenges~\cite{eghbal-zadehCPJKUSubmissionsDCASE20162016,DorferDCASE2018task1,SakashitaDCASE2018task1}  are still based on VGG variants. We will show in Section~\ref{sec_exp} how we adapted ResNet-28 (with 28 layers) to finally outperform VGG models.
% Resnet consist of 

\subsubsection{DenseNet}
DenseNet~\cite{huangDenselyConnectedConvolutional2017} 
 solves the vanishing gradient problem by concatenating the outputs of all previous convolution layers on the channels dimension. DenseNet variants achieves state-of-the-art results on computer vision tasks.
% \gerhard{one or two references?} \khaled{I omited those for space}
% \gerhard{Is DenseNet or ResNet the more current "state of the art"? } \khaled{ they all achieves state-of-the-art, https://github.com/arunpatala/cifarSOTA (with small differences in performance, but usually they're all called sota). the better variants are shake-shake and resnext. we tried this hypothesis on both shake-shake and resnext and it works but I didn't find place here to add more architectures}

\subsection{Setup}
For all experiments, we set up our training framework following the current state of the art~\cite{DorferDCASE2018task1,SakashitaDCASE2018task1}. All the experiments used the same setup, to ensure a fair comparison.

\subsubsection{Data Preparation}
The input is down-sampled to 22.05 kHz and subjected to a Short Time Fourier
Transform (STFT) with a window size of 2048 and 25\% overlap, followed by a Mel-scaled filter bank on perceptually weighted spectrograms. That results in 256 Mel frequency bins and around 43 frames per second. The input frames are normalized %using the training set mean and standard deviation.
to zero-mean and unit variance according to the training set.
 \subsubsection{Optimizing}
The Adam optimizer~\cite{kingmaAdamMethodStochastic2014} is used for a total of 350 epochs, with a starting learning rate of $1 \times 10^{-4}$. The learning rate decays linearly from epoch 50 until 250 where it reaches $5 \times 10^{-6}$. Then we train for another 100 epochs with the minimum learning rate $5 \times 10^{-6}$. 

 \subsubsection{Data Augmentation}
We use \textit{mix-up}~\cite{zhangMixupEmpiricalRisk2017} since it was shown to improve the generalization of the models and to help in preventing overfitting.

 \subsubsection{Obtaining Results}
The models are evaluated on the test set after 350 epochs of training. By using mix-up and batch normalization in the architectures, we ensure that models are not overfitting on the training set.
%Moreover we observe the loss/accuracy curves for the reported experiments \improvement{maybe different way to say this}.\gerhard{Yes, please. What do you mean to say? Did you just "look at" the curves or did you make any decisions based on "observing" the curves?} \khaled{I meant here that the testing loss never has an up-trend , there is some variance but not the classical overfitting. should we just omit this sentense?}
Each experiment is repeated 3 times for Table~\ref{tab_vision_vs_rf} and 6 times for Figure~\ref{fig_dcase16}.
%\gerhard{how many times?} \khaled{at least three times. at first I was repeating 6 times (most of the curves) but when adding new architectures and datasets that became too much. so I started repeating 3 time only. should I rerun more?, or say "3 times at least  }; 
We report the mean and the standard deviation of these runs.

%% file: sec_res.tex
%!TEX root = main.tex
\subsection{Modifying the ``Vision Architectures''}
In this section we report on a series of experiment whose
purpose is to investigate the effect of the RF on the performance of a CNN architecture. In particular, we investigate how and
to what extent we can push a CNN's generalization by calibrating its RF.
We do this by evaluating modified ResNets and DenseNets on several benchmark datasets from the world
of acoustic scene classification.
\label{subsection:modifyvisionarcs}
% \gerhard{Maybe we can find a better subsection title ...}

Table~\ref{tab_vision_vs_rf} shows the performance of the ``vision architectures'' ResNet and DenseNet before and after adjusting the RF to match the VGG-like architecture proposed by Dorfer et al.~\cite{DorferDCASE2018task1}.
We modified the deep architectures from Section~\ref{sec_archs} to obtain similar RFs as~\cite{DorferDCASE2018task1}, as described below.
These straight-forward modifications boost the performance across all architecture and datasets. The adopted models outperform both the VGG baseline as well as their vision-inspired predecessors. 
\subsubsection{ResNet}
ResNet is usually deeper than VGG, and thus has a larger receptive field. There are many ways to modify ResNet to have a smaller RF; we chose the following:

\begin{table}[t!]
\caption{Architectures performance on DCASE datasets}
\begin{center}

\begin{tabular}{rccc}

         & DCASE16            & DCASE17            & DCASE18      \vspace{.03in}      \\ 
%\\
&\multicolumn{3}{c}{Loss}                                              \\ \hline
DenseNet~\cite{huangDenselyConnectedConvolutional2017}& $ 0.58 \pm 0.02 $  & $ 1.31 \pm 0.17 $  & $ 0.92 \pm 0.04 $  \\ 
ResNet~\cite{heDeepResidualLearning2016}& $ 0.67 \pm 0.06 $  & $ 1.19 \pm 0.06 $  & $ 1.02 \pm 0.05 $  \\ 
VGG~\cite{simonyanVeryDeepConvolutional2014,DorferDCASE2018task1}        & $ 0.69 \pm 0.05 $  & $ 1.05 \pm 0.04 $  & $ 0.83 \pm 0.03 $  \\ \hline
RN1      & $ 0.57 \pm 0.02 $  & $ 0.94 \pm 0.09 $  & $ 0.67 \pm 0.03 $  \\ 
RN2      & $ 0.50 \pm 0.02 $  & $ 0.91 \pm 0.04 $  & $ 0.71 \pm 0.03 $  \\ 
RN3      & $ 0.53 \pm 0.04 $  & $ 0.89 \pm 0.02 $  & $ 0.67 \pm 0.00 $  \\ 
DN1      & $ 0.51 \pm 0.01 $  & $ 0.90 \pm 0.04 $  & $ 0.72 \pm 0.06 $  
\vspace{.03in}\\ 
%\\
&\multicolumn{3}{c}{Accuracy}                                          \\ \hline
DenseNet~\cite{huangDenselyConnectedConvolutional2017} & $ 83.68 \pm 0.90 $ & $ 63.48 \pm 4.96 $ & $ 71.55 \pm 0.85 $ \\ 
ResNet~\cite{heDeepResidualLearning2016}   & $ 83.17 \pm 0.91 $ & $ 67.19 \pm 1.72 $ & $ 71.05 \pm 0.87 $ \\ 
VGG~\cite{simonyanVeryDeepConvolutional2014,DorferDCASE2018task1}      & $ 82.99 \pm 0.90 $ & $ 67.90 \pm 1.31 $ & $ 74.56 \pm 1.01 $ \\ \hline
RN1      & $ 85.98 \pm 1.32 $ & $ 71.11 \pm 1.19 $ & $ 77.34 \pm 1.53 $ \\ 
RN2      & $ 87.09 \pm 0.53 $ & $ 72.41 \pm 0.96 $ & $ 75.71 \pm 0.70 $ \\ 
RN3      & $ 86.51 \pm 1.05 $ & $ 71.74 \pm 0.85 $ & $ 77.61 \pm 0.22 $ \\ 
DN1      & $ 86.07 \pm 0.65 $ & $ 72.24 \pm 1.00 $ & $ 76.39 \pm 0.14 $ \\ 
\end{tabular}
\end{center}
\label{tab_vision_vs_rf}
\end{table}

\begin{itemize}
\item \textbf{Making ResNet shallower} (\textit{RN1} in Table~\ref{tab_resnet_configs})
We removed tailing layers from ResNet after the RF of~\cite{DorferDCASE2018task1} is reached. The resulting network looks similar to VGG, but with residual connections; the network consists of 5 residual blocks of two convolutional layers each. The resulting network has a total of $3,258,772$ trainable parameters, since we did not change the number of channels in the new network.

\item \textbf{Changing filter sizes} (\textit{RN2} in Table~\ref{tab_resnet_configs})
Instead of removing layers from ResNet, We changed some filter sizes from $3\times3$ to $1\times1$. Network \textit{RN2} has $12$ residual blocks of two convolutional layers each. $7$ residual blocks have $3\times 3$ filters in the first convolutional layer and $1 \times 1$ and the second layer; the rest of the residual blocks consist only of $1 \times 1$ convolutions. In other words, we kept the depth of ResNet but changed many filter sizes to $1 \times 1$. \textit{RN2} has a total number of parameters of $6,053,780$. 

% compare to vgg has 6499688 params

\item \textbf{Changing the sizes of the tailing filters}  \textit{RN3} is similar to \textit{RN1}, except that we do not delete the tailing layers but instead only change their filter sizes to $1 \times 1$.
% params 3959198

\end{itemize}
\begin{table}[t!]
\caption{Modified ResNet architectures}
\begin{center}
\begin{tabular}{|c|c|c|c|}
\hline
\textbf{RB Number}&\multicolumn{3}{|c|}{\textbf{RB Config}} \\
\cline{2-4} 
\textbf{} & \textbf{\textit{RN1}}& \textbf{\textit{RN2}}& \textbf{\textit{RN3}} \\
\hline
&\multicolumn{3}{|c|}{Input $ 5 \times 5$ stride=$2$ }
\\
\hline

1& $3 \times 3$, $ 1 \times 1$, P  & $3 \times 3$, $ 1 \times 1$, P&$3 \times 3$, $ 1 \times 1$, P\\
2& $3 \times 3$, $ 3 \times 3$, P  & $3 \times 3$, $ 1 \times 1$  &$3 \times 3$, $ 3 \times 3$, P  \\
3& $3 \times 3$, $ 3 \times 3$,  & $3 \times 3$, $ 1 \times 1$ &$3 \times 3$, $ 3 \times 3$  \\
4&   & $1 \times 1$, $ 1 \times 1$, P &$3 \times 3$, $ 3 \times 3$, P   \\
5& $3 \times 3$, $ 3 \times 3$, P& $3 \times 3$, $ 1 \times 1$ &$3 \times 3$, $ 1 \times 1$  \\
6& & $3 \times 3$, $ 1 \times 1$ &$ 1 \times 1$, $ 1 \times 1$  \\
7& & $3 \times 3$, $ 1 \times 1$ &$ 1 \times 1$, $ 1 \times 1$ \\
8& & $3 \times 3$, $ 1 \times 1$, P &$ 1 \times 1$, $ 1 \times 1$  \\
9& $3 \times 3$, $ 1 \times 1$ & $3 \times 3$, $ 1 \times 1$ &$ 1 \times 1$, $ 1 \times 1$  \\
10& &$ 1 \times 1$, $ 1 \times 1$ &$ 1 \times 1$, $ 1 \times 1$  \\
11& & $ 1 \times 1$, $ 1 \times 1$ &$ 1 \times 1$, $ 1 \times 1$  \\
12& & $ 1 \times 1$, $ 1 \times 1$ &$ 1 \times 1$, $ 1 \times 1$  \\
\hline
\multicolumn{4}{l}{RB: Residual Block, P: $ 2 \times 2$ max pooling after the block. }\\
\multicolumn{4}{l}{RB number 1-4 have 128 channels, RB number 5-8 have 256 channels,}\\
\multicolumn{4}{l}{RB number 9-12 have 512 channels.}
\end{tabular}
\label{tab_resnet_configs}
\end{center}
\end{table}
 
\subsubsection{DenseNet} 
Unlike ResNet and VGG, estimating the effective RF of DenseNet is non-trivial. It cannot be inferred directly from the maximum possible RF  \eqref{eq_max_rf} because of the dense connections~\cite{huangDenselyConnectedConvolutional2017}. Each convolutional layer projects its input into feature maps with a small number of channels (the \textit{`growth rate'} in ~\cite{huangDenselyConnectedConvolutional2017}). This will increase the maximum RF but will have a small effect since only few feature maps have this RF. We increased the growth rate to 128 and reduced the network depth in order for its maximum RF to match Dorfer et al.~\cite{DorferDCASE2018task1}. The resulting network \textit{DN1} has $5,269,902$ parameters.

% original densnet params 4844894

% \begin{table}[htbp]
% \caption{Table Type Styles}
% \begin{center}
% \begin{tabular}{|c|c|c|c|c|c|c|}
% \hline
% \textbf{Network}&\multicolumn{6}{|c|}{\textbf{Dataset}} \\
% \cline{2-7} 
% \textbf{} & \multicolumn{2}{|c|}{\textbf{\textit{DCASE16}}}& \multicolumn{2}{|c|}{\textbf{\textit{DCASE17}}}& \multicolumn{2}{|c|}{\textbf{\textit{DCASE18}} }\\
% \hline
%  & loss & acc  & loss & acc  & loss & acc \\\hline
% DenseNet& $.56 \pm ?$  & $.83 \pm ?$ & $1.38 \pm?$ & $.624 \pm ?$ & $.88 \pm ?$ & $.72 \pm ?$ \\
% ResNet&  & & &  & &  \\
% VGG&  & & &  & &  \\
% \hline
% \textit{ResNet\_1}& $.58 \pm .02$&$.86 \pm .01$ & $.94 \pm .06$ &$.69 \pm .01$  & $.70 \pm .02$ & $.76 \pm .01$ \\
% \textit{ResNet\_2}&  & & &  & &  \\
% \textit{ResNet\_3}&  & &  &  & & \\
% \textit{DenseNet\_1}&$.51 \pm ?$  & $.853 \pm ?$ & $.88 \pm ?$ & $.72 \pm ?$ & $.69 \pm ?$ & $.76 \pm ?$ \\
% \hline

% \multicolumn{4}{l}{$^{\mathrm{a}}$Sample of a Table footnote.}
% \end{tabular}
% \label{tab_vision_vs_rf}
% \end{center}
% \end{table}

%%%%%%%%%%%%
% @TODO UPDATE RN2 on DCASE2019 LAST RUN RK4

%%%%%%%%%%%%

\subsection{The Effect of Systematic Changes of the RF}
\label{subsec:changeresnet}

\begin{figure*}[t!]
\centerline{\includegraphics[]{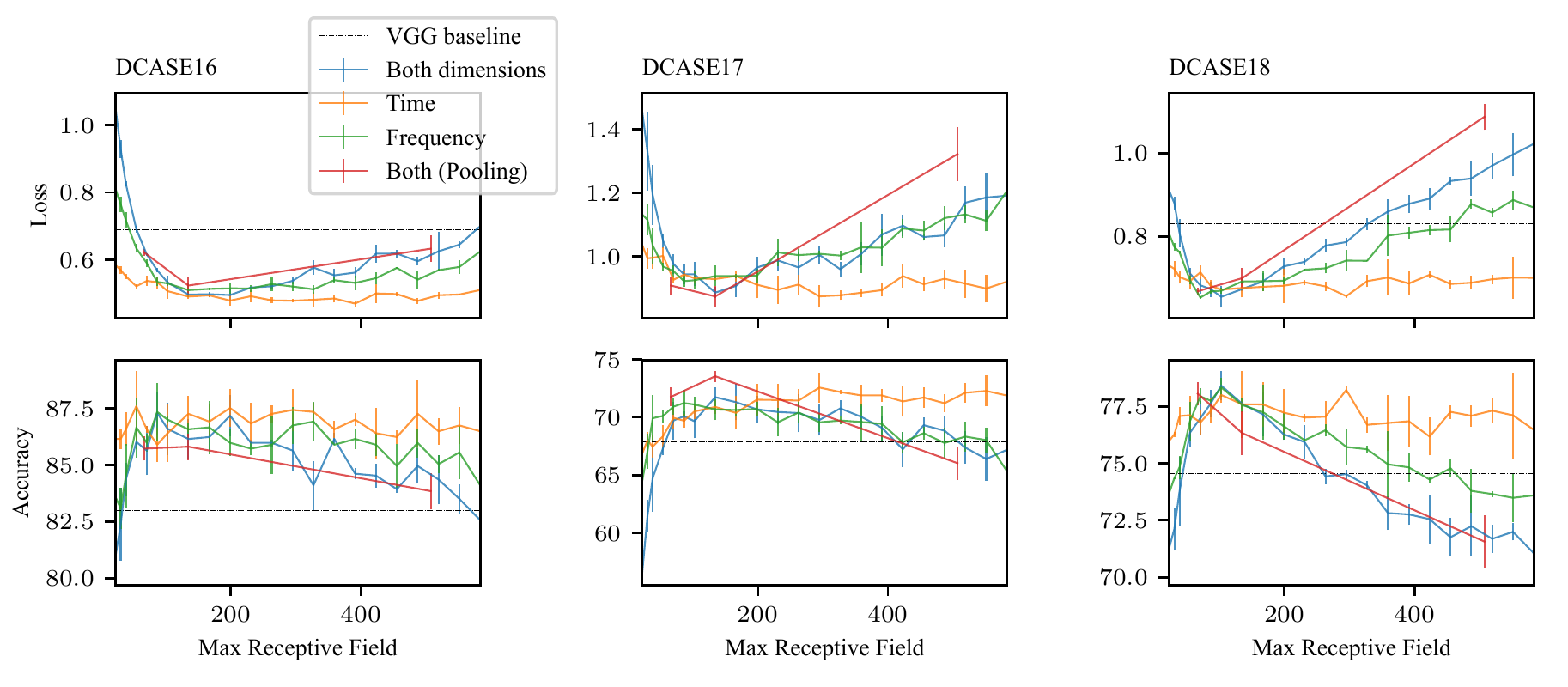}}
\caption{The effects of systematic changes to the receptive field of ResNet (averages and std.~deviations over 6 runs).
The dashed horizontal line is the loss/accuracy achieved by
the current state-of-the-art VGG architecture in this field~\cite{DorferDCASE2018task1,DorferDCASE2018task2}.}
% (Section~\ref{subsec:changeresnet}). }
\label{fig_dcase16}
\end{figure*}

For more detailed insight, Fig.~\ref{fig_dcase16} shows the testing loss and accuracy for various configurations of our ResNet, modified
% We modified the ResNet architecture % (Section~\ref{sec_res})
by systematically varying the network filters or the pooling:

\subsubsection{Changing filters in both dimensions} 
 Starting from the last layer of a network with all $3 \times 3$ filters, we modify the filters from $3 \times 3$ to $1 \times  1$. This will affect the RF on both dimensions, and results in networks with maximum RFs (Section~\ref{sec_rf}) ranging from $23\times23$  to $583 \times 583$ pixels.

\subsubsection{Changing filters in the time dimension only} 
 Here, we fix the RF on the frequency dimension to the one in the Dorfer et al. model~\cite{DorferDCASE2018task1} (i.e., 135 pixels). We only change the filter sizes in the time dimension. This results in networks with maximum RFs ranging from $23\times135$  to $583 \times 135$ pixels.

\subsubsection{Changing filters in the frequency dimension} 
The procedure is analogous to the previous one, but using the frequency dimension instead of time. This results in networks with maximum RFs ranging from $135\times23$  to $ 135 \times 583$ pixels.

\subsubsection{Changing pooling layers} 
\label{subsec:change_pooling}
Here, we add or remove pooling layers. 
Pooling layers have a stride of 2; adding one thus doubles accumulative stride (see Eq.~\eqref{eq_max_rf}) in all layers that follow the pooling. 
% Pooling layers also increase the training speed since they downsample the spatial dimensions.
%maybe the following sentence is not important
% Therefore, making fine steps in the RF size or achieving very low RF using pooling was not feasible and out of the scope of this work.
% \improvement{maybe a better way of saying this}. 
% \gerhard{Best solution: don't say it at all.}
Pooling layers do not affect the number of parameters of the network, hence we use them to demonstrate the effect of the RF without changing the number of parameters.
However, adding a pooling layer downsamples the spatial dimensions by the provided factor and the RF of all subsequent layers will be affected by this change.
Because of these architectural limitations, we only provide the feasible steps in this experiment.
% adding/removing pooling layers does not allow us to finely control the RF in small steps.
Figure~\ref{fig_dcase16} shows the results of networks with RFs of $ 67 \times 67$, $ 135 \times 135$, and $507 \times 507$.

% \begin{figure}[htbp]
% \centerline{\includegraphics[]{figs/dcase16_128c2_s1.pdf}}
% \caption{DCASE 16 The effective Receptive of 3 different architectures explained in Section~\ref{sec_exp}.}
% \label{fig_dcase16}
% \end{figure}

% \begin{figure}[htbp]
% \centerline{\includegraphics[]{figs/dcase17_128c2.pdf}}
% \caption{DCASE 17 The effective Receptive of 3 different architectures explained in Section~\ref{sec_exp}.}
% \label{fig_dcase17}
% \end{figure}

% \begin{figure}[htbp]
% \centerline{\includegraphics[]{figs/dcase18_128c2.pdf}}
% \caption{DCASE 18 The effective Receptive of 3 different architectures explained in Section~\ref{sec_exp}.}
% \label{fig_dcase18}
% \end{figure}

% \begin{figure}[htbp]
% \centerline{\input{figs/dcase16_128c.pgf}}
% \caption{The effective Receptive of 3 different architectures explained in Section~\ref{sec_exp}.}
% \label{fig_dcase16}
% \end{figure}

%% file: sec_disc.tex
%!TEX root = main.tex
\subsection{The Effect of the Receptive Field}

Figure~\ref{fig_dcase16} shows the effect of systematic changes to the RF, on three ASC datasets. 
It can be seen that there is a certain range of RF sizes that permit
the CNN to generalize and predict well. On the one hand, RFs that are too small impair the CNN's performance. Our intuition is that output neurons do not collect enough information to make the optimal decision and thus underfit even the training data. On the other hand, when the RF grows larger than this range -- usually in the case of the CNN growing deeper -- the CNN overfits the training data and fails to generalize to new samples. We observed this in the training loss (not reported) which decreases the bigger the RF is, while the performance deteriorates. This effect is more clear in DCASE17 and DCASE18 compared to DCASE16. We explain this with the characteristics of the datasets: in DCASE16, the test set has a similar distribution to the training set, while in DCASE17 and DCASE18 the impact of a (non-)optimal RF is magnified because of the distribution mismatch between the test and training sets.

\subsection{The Effect of Changing the RF over one Dimension }

The plots also show that an optimal RF over the frequency dimension makes the effect of increasing the RF over the time dimension surprisingly minimal, provided that it is large enough to capture the necessary information. A possible reason may be that the training datasets have enough variance in the temporal dimension of the learned features so that the model does not overfit on long-term temporal dependencies in the data. However, the opposite happens when shrinking/extending the RF on the frequency dimension, despite an optimal RF over time. Seeing the whole frequency range seems to lead the CNN to overfit on the training data's acoustic events, which may differ in some frequency bins in the test recordings.

\subsection{The Influence of the Number of Parameters}
\label{subsec:param_num}

Finally, we see that using different CNNs with the
same number of parameters but with different RFs (Section~\ref{subsec:change_pooling}) results in a performance that correlates with the results of the previous section. This indicates that the effect on overfitting of a larger RF is more significant than the number of parameters for a given architecture. For example, RN1, RN2 and RN3 have different numbers of parameters but the same RF; they perform similarly compared to other architectures.

%% file: sec_con.tex
%!TEX root = main.tex

We investigated the relation between CNNs' RFs over the input spectrograms and their generalization on unseen samples, for the acoustic scene classification task. We showed that a large RF especially over the frequency dimension pushes CNNs to overfit, while a smaller than necessary RF forces a CNN to underfit the data and prevents it from learning decisive features. 
Although many factors contribute to a CNN's tendency to generalize, we show that for a specific training setup and network architecture, tuning the RF of the model is a crucial factor for its performance.
Following these guidelines, we managed to tune state-of-the-art vision architectures that so far failed to perform well on ASC, to match and outperform current
popular and
state-of-the-art CNNs on this task. 
% ASC CNNs, by addressing the problems of such models when adapted from a vision to acoustic context.